# A new Sparse Auto-encoder based Framework using Grey Wolf Optimizer for Data Classification Problem

Ahmad M. Karim[1,*]

Computer Engineering, Istanbul Gedik University, Istanbul, Turkey

Corresponding Author: Ahmad.karim@gedik.edu.tr

**Abstract**

One of the most important properties of deep auto-encoders (DAEs) is their capability to extract high level features from row data. Hence, especially recently, the autoencoders are preferred to be used in various classification problems such as image and voice recognition, computer security, medical data analysis, etc. Despite, its popularity and high performance, the training phase of autoencoders is still a challenging task, involving to select best parameters that let the model to approach optimal results. Different training approaches are applied to train sparse autoencoders. Previous studies and preliminary experiments reveal that those approaches may present remarkable results in same problems but also disappointing results can be obtained in other complex problems. Metaheuristic algorithms have emerged over the last two decades and are becoming an essential part of contemporary optimization techniques. Gray wolf optimization (GWO) is one of the current of those algorithms and is applied to train sparse auto-encoders for this study. This model is validated by employing several popular Gene expression databases. Results are compared with previous state-of-the art methods studied with the same data sets and also are compared with other popular metaheuristic algorithms, namely, Genetic Algorithms (GA), Particle Swarm Optimization (PSO) and Artificial Bee Colony (ABC). Results reveal that the performance of the trained model using GWO outperforms on both conventional models and models trained with most popular metaheuristic algorithms.

## 1. Introduction

Gene expression is primarily the name of the process used by the synthesis of functional gene product. This process involves two stages namely, transcription and translation. While the first stage is responsible from the process of RNA synthesis, the translation stage, on the other hand, employs mRNA for direct protein synthesis and also involves further post-translational processing of the corresponding molecules. Within the last decade, microarray technologies are adapted measure the expression levels of large numbers of genes, which, however, are not accurate sufficient to prophesy. Especially, the recent increase in machine learning's presence at the solution of different problems encourage researchers to adapt those algorithms to classify and analyze gene expressions. Conventional machine learning approaches have encountered with classical problems during learning procedure, namely, noisy data, high dimensionality and small amount of data for training. Accordingly, those techniques cannot achieve satisfactory results when adapted to the corresponding field [1, 2, 3].

Deep learning (DL) is a new field of machine learning which involves complex nonparametric mathematical models, adapted by the biological nervous system. Since the rise of DL it has been applied to various classification problems such as face recognition, medical imaging, voice analysis, and natural language process and swarm intelligence. Results reveal that DL architectures positively affected learning procedures [4, 5, 32]. Sparse auto-encoder (SAE) is a recent but rapidly growing DL technique, which has been already applied effectively to several fields [6, 7]. Overall, the approach that offers learning for an auto-encoder is known as trainer. A trainer is accountable for training auto-encoders to extract the best features for inputs of unlabeled dataset.

During the training phase of a SAEs, first the auto-encoders are trained by a set of examples namely, training examples. The trainer then changes the structural parameters of the auto-encoders within each training step so as to enhance the overall performance of the architecture [8]. Once, the training stage is completed according to the given stopping criteria, the auto-encoder is prepared to employ to the test data, if the results are not satisfactory, the model is retrained by both considering the structure of the model and modifying parameter values.

Gradient-based methods tend to be stuck in local minima, and are highly dependent on the initial parameters [9, 10]. Consequently, several different metaheuristic algorithms are recently preferred to train deep neural network in several studies such as PSO in [11], hybrid artificial bee colony in [12], and water wave optimization algorithm to train traditional neural network and deep neural network in [13], and genetic algorithm to train auto-encoder in [14]. On the other hand, results represent that they also offer several weaknesses like high dependence to the initial solution, causing late convergence, as well as local optima trap, a popular case in nonlinear problems. In a recent study, an optimized version of Binary Grey Wolf Algorithm is proposed to select optimal features for classifying acute leukemia [33]. Another similar study also proposes a framework involving improved GWO and Kernel Extreme Learning Machine algorithms to be used by medical datasets [34]. Besides, a recent study also combines the sine cosine algorithm (SCO) and GWO. The aim of this hybrid model is to provide a better performance on complex benchmark problems of engineering [35]. Another recently published paper also presents a model for cancer classification by integrating GWO algorithm into conventional machine learning classifiers [36]. Some other relevant studies can be seen in [37, 38].

Despite the recent popularity and high performance of autoencoders in various machine learning problems. The training phase of autoencoders is still a challenging task. The main motivation lies behind this study to propose a new training approach to select appropriate parameters, allowing the DAE model to approximate best results. Accordingly, GWO algorithm, performs superior to the other metaheuristic algorithms, is adapted to train sparse auto-encoders for this study. This model is validated by employing several popular benchmark databases. The experimental results are compared with state-of-the art machine learning methods using the same datasets, and also are compared with leading intelligent optimization algorithms.

This study, essentially, presents a new framework using the stacked sparse autoencoders (SSAEs) architecture involving multiple layers of SAEs. Aforementioned, one of the main contributions of this study is to employ Gray Wolf Optimization (GWO) to train SAEs. The proposed framework is validated by using popular and comprehensive gene expression datasets. The framework is compared with previous studies using the same datasets, as well as comprehensive metaheuristic methods are employed to train SAEs so as to compare them with GWO using the same datasets. Confidently, the proposed framework outperforms previous researches based on the same datasets. Besides, GWO outperforms three other popular metaheuristic algorithms used in similar problems. Overall, the structure of paper consists of the current literature, proposed method, experiments and conclusion.

## 2. Literature Review

This section presents several comprehensive and milestone studies associated with gene expression data classification. V. Nandagopal et al. (2019) proposes a fuzzy based logistic regression for gene data feature selection to diagnosis breast cancer disease. The dataset used for experiments consist of 116 examples, they claim that the proposed Logistic Regression (LLR) model archives 94.05% accuracy within the test data [15]. Ping He et al. (2019) proposes a sparse learning model for gene expression data classification, namely, "Group K-Singular Value Decomposition", the optimum thesaurus and sparse representation obtained from training data. Authors claim that their method presents satisfactory results when it is validated with popular datasets, including Leukemia, DLBCL, SRBCT, Brain, Prostate, Lung, and Amazon datasets [16]. In [17] Maniruzzaman et al. (2019) extract gene set to recognize cancer disease based on 10 different classifiers. Results reveal that the combination of "WCSRS" test with "RF-based" yields 99.81% accuracy, considered the highest results among all techniques studied in the corresponding field. In a recent study, a new version of k-nearest neighbour (KNN) classification method for gene expression data is proposed, namely "Modified k-nearest neighbour (MKNN)". According to the situation, two versions of the method are employed, namely "largest modified KNN (LMKNN)" and "smallest modified KNN (SMKNN)". Both are presented to optimize the performance of KNN. This modified version of the algorithms are compared with popular supervised classifiers such as KNN, Support Vector Machines (SVM), and weighted KNN. Authors state that the proposed method outperforms given classifiers in overall accuracy and computational time [18]. Huijuan Lu et al presents a new feature extraction method, combining information maximization and the adaptive genetic algorithm, are also presented [19]. The presented method both removes redundancies and reduces the dimension of data so as to increase the classification accuracy. Yuanyu He et al. (2019) proposes a new method called "imRelief". The proposed method runs successfully with imbalanced high-dimensional medical data. Experimental results reveal the efficiency of "imRelief" feature weighting and subset specification approaches with microarray data [20]. In another study, a distributed feature selection method using Symmetrical Uncertainty with Multilayer Perceptron is presented [21]. While, the Symmetrical Uncertainty is applied to select important features, the Multi-Layer Perceptron is employed to classify the selected features. This method is confirmed by using a well-known dataset, having 7 dimensions, and only achieves 58% success rate [21]. Finally, a hybrid method, combining Information Gain and SVM techniques for gene classification problem, is studied and published in [22]. According to the shared results, "IG-SVM" approach carries out 90.32% accuracy to classify colon cancer dataset. Chen et al presents a multi-layer autoscored architecture involving annotated gene set data. The results reveal that the model is capable of preserving critical features of gene expression data during the dimension reduction process [39]. Jia Wen et al, on the other hand, designs a deep auto-encoder (DAE) model to forecast gene expression from SNP genotypes. The model is evaluated by using a benchmark dataset and the superiority of the model is demonstrated [40]. Peng et al introduces an interesting study which basically combines gene ontology with DAEs so as to cluster single cell RNA-Seq data in a more efficient way. Results proves the advantage of the proposed method over the state-of-the-art methods in terms of dimension reduction [41]. Some of those state-of-the-art literature, employing the same datasets defined in this study, are compared with the proposed model. Some other relevant studies are presented in [42, 45, 51]. Table 2 presents the comparison results between the proposed architecture and some of these relevant studies.

## 3. Methodology

### 3.1 Sparse auto-encoder

Autoencoders (AEs) are a kind of feedforward neural networks, compressing the input data into a lower-dimensional code and then reconstruct the output from this demonstration. AEs are classified as an unsupervised learning technique. Accordingly, they do not need explicit labels for training process. Sparse Autoencoders (SAEs) are a specific type AEs that relies on regularization. In order to regularize AEs, a sparsity constraints, a penalty term for the loss function is added. This allows the model to learn better representations of input than conventional AEs. It should be noted that when SAEs are able to reduce the dimension of the data, they can also discover interesting structure in the input data. This motivates researchers to employ SAEs as feature extractors, followed by a classifier to be trained in a supervised manner. Overall, the first SAE layer is used to obtain feature vector from the input, which is then used as the input for the following SAE layer. This process continues until the training is completed in an unsupervised manner. Afterwards, backpropagation algorithm (BP) is employed to minimize the cost function and enhance the weights by considering the labelled training set to achieve classification in a supervised manner. This model is called as the stacked sparse auto-encoder (SSAE) and detailed in the following paragraph. The stacked sparse auto-encoders (SSAEs) concept essentially refers a neural network having numerous auto-encoders. Each denotes a layer and is trained by using an unsupervised methodology with unlabeled data [23]. The input layer of each auto-encoder is linked to the output layer of the previous one. The training phase principally aims to estimate the optimal parameters of an algorithm by considering different algorithms. Those algorithms mainly aim to decrease the divergence between input "$x$" and output "$\dot{x}$". The corresponding steps taken place between those layers is denoted as follows:

$$\dot{x} = f(x) \tag{1}$$

$$n_1^{(1)} = M_f(w_{11}^{(1)} x_1 + \cdots w_{15}^{(1)} x_5 + b_1^{(1)}) \tag{2}$$

$$n_i^{(1)} = M_f(w_{i1}^{(1)} x_1 + \cdots w_{i5}^{(1)} x_5 + b_i^{(1)}) \tag{3}$$

Here, M symbolizes activation function, namely, "sigmoid logistic function".

The ultimate mathematical model is explained in Eq. (4):

$$n_{w,b}(x) = M_f(w_{11}^{(2)} n_1^{(2)} + \cdots w_{15}^2 n_5 + \cdots + b_1^{(2)}) \qquad (4)$$

The input $x$ and output $\dot{x}$ divergence is signified by using a cost function. Several algorithms and approaches may be employed to calculate the optimum parameters of the network. It should be noted that ReLU activation function is preferred especially by CNNs due to its computational efficiency. However, the proposed model does not need lots of computational power due to the scale of our training sets. Besides, ReLU function can also encounter with Dead Neuron and Bias Shift problems. Accordingly it is preferred to employ sigmoid function, very popular activation function for neural networks.

The proposed SSAE model consists of two stacked auto-encoders layers for feature extraction and a SoftMax layer for the classification process. Auto-encoders attempt to extract the sensitive and high-level features from the input data "X". The main intention of using more than one auto-encoder is to reduce number of features steadily. That is because reducing the number of features rapidly by relying on a single auto-encoder can cause to miss substantial features during the reducing process, which certainly decreases the overall performance of the system. The cost function of a sparse auto-encoders is represented by using an adjusted mean square error function as shown in Eq. (5).

$$E = \frac{1}{N} \sum_{n=1}^{N} \sum_{k=1}^{K} (x_{kn} - \hat{x}_{kn})^2 + \lambda * \Omega_{weights} + \beta * \Omega_{sparsity}$$

(5)

Here, the error is represented by "$E$", "$N$" signifies the total number of training samples and "$K$" symbolizes the total attribute numbers of each sample. The input features are represented by "x". While the "reconstructed features" are represented by "$\hat{x}$", $\lambda$ represents the coefficient for the "L2 Weight Regularization", $\beta$ illustrates the coefficient for "Sparsity Regularization", and finally $\Omega_{weights}$ represents the L2 Weight Regularization, which can be obtained as follow:

$$\Omega_{weight} = \frac{1}{2} \sum_{l}^{L} \sum_{n}^{N} \sum_{k}^{K} w_{ji}^{(l)^2}$$

(6)

Where, L represents the hidden layer numbers, n represents observations (training samples), and k represents the attribute number. Finally, $\Omega_{sparsity}$ is the "Sparsity Regularization" parameter regulates the influence of sparsity for faster optimization and evaluation of the proposed model.

One of the leading sparsity regularization terms is the "Kullback-Leibler (KL) divergence", which basically measures the differences between two distributions. Corresponding equation is given in (7).

$$\Omega_{sparsity} = \sum_{i=1}^{D} KL(\rho||\hat{\rho}_i) = \sum_{i=1}^{D} \rho \log(\rho||\hat{\rho}_i) + (1-\rho) \log\left(\frac{1-\rho}{1-\hat{\rho}_i}\right) \quad (7)$$

Here, the desired output activation value of each neuron is symbolized by $\rho$, $\hat{\rho}_i$ represents the average output activation of a neuron "$i$", and "$KL$" is the function, calculating the disparity between two distributions throughout the same data, whereas "D" denotes the number of neuron in hidden layers. Besides, the features generating the minimum cost illustrated in Eq. (5). The details of mathematical model, involving SSAE are presented in [24].

### 3.2 Grey Wolf Optimizer (GWO)

The Grey Wolf Optimizer (GWO) is a newly presented swarm-based metaheuristic algorithm. This algorithm impersonator the common guidance and hunting behavior of wolves in wildlife. Grey wolves enclose their prey during the hunt. This behavior 'D' is defined in Equations 8 and 9. According to the mathematical model, the population is separated into four sets, namely, "alpha" ($\alpha$), "beta ($\beta$)", delta ($\delta$), and "omega ($\omega$)". The fittest wolves, are given as $\alpha$, $\beta$, and $\delta$, are responsible to lead other wolves ($\omega$) to auspicious areas of the exploration space. Within the optimization procedures, the wolves modify their positions by considering "$\alpha$", "$\beta$", or "$\delta$" parameters, whereas as $\vec{A}$ and $\vec{C}$ refers coefficient vectors, illustrated in Equations 8 and 9.

$$\vec{D} = |\vec{C} \cdot \vec{X_p}(t) - \vec{X}(t)| \quad (8)$$
$$\vec{X}(t+1) = \vec{X_p}(t) - \vec{A} \cdot \vec{D} \quad (9)$$

Here, t denotes the recent iteration, $\vec{A} = 2b \cdot \vec{t_1}b$, $\vec{C} = 2 \cdot \vec{t_2}$, $\vec{X_p}$ represents the position of the prey, whereas $X$ indicates the position of the wolf. This is a linearly decreased value between 1 and 2, while $t_1$ and $t_2$ represent random vectors defined in [0,1]. In GWO algorithm, the $a$, $\beta$, $\delta$ represent the prey optimum position [25, 26]. With the intention of mathematically simulating the hunting behavior of wolves, it should be expected that the alpha (the finest contender solution) beta, and delta have superior information about the possible position of the prey. According to which positions of the best three agents are calculated as follows:

$$\vec{D_\alpha} = |\vec{C_1} \cdot \vec{X_\alpha} - \vec{X}| \quad (10)$$
$$\vec{D_\beta} = |\vec{C_2} \cdot \vec{X_\beta} - \vec{X}| \quad (11)$$
$$\vec{D_\delta} = |\vec{C_2} \cdot \vec{X_\delta} - \vec{X}| \quad (12)$$

Here, positions of alpha, beta and delta are represented by $\vec{X}_\alpha, \vec{X}_\beta$ and $\vec{X}_\delta$ respectively. While, $\vec{C1}, \vec{C2}$ and $\vec{C3}$ represent arbitrary vectors, X denotes the position of the present solution. Equations (10-12) estimate the rough distance between the existing solution and $\vec{X}_\alpha, \vec{X}_\beta$ and $\vec{X}_\delta$ correspondingly. Other search agents (comprising the omegas) are expected to modify their positions in relation to the positon of the finest search agent. After obtaining the distances, the final position of the existing solution is calculated as shown below. Equations from 13 to 15 illustrate the best solutions with respect to the alpha" ($\alpha$), "beta ($\beta$)" and delta ($\delta$). Other search agent (comprising the omegas) modifies its position accordingly, which can be seen in Eq. 16.

$$\vec{X1} = \vec{X_\alpha} \cdot \vec{R1} - (\vec{D_\alpha}) \tag{13}$$
$$\vec{X2} = \vec{X_\beta} \cdot \vec{R2} - (\vec{D_\beta}) \tag{14}$$
$$\vec{X3} = \vec{X_\delta} \cdot R_3 - (\vec{D\delta}) \tag{15}$$
$$\vec{X}(t+1) = \frac{(\vec{X_1} + \vec{X_2} + \vec{X_3})}{3} \tag{16}$$

Here, while $\vec{R1}, \vec{R2}$ and $\vec{R3}$ vectors refer random vectors, t refers the iteration number. Equations from 11 to 13 describe the period size of the w wolves to α, β, and δ parameters correspondingly. Eq. 16 calculates the last position of the wolves (w) by considering Equations 13,14 and 15. The GWO algorithms steps is given below. The details of GWO algorithm can be seen in [27].

*Steps of the Algorithm GWO (w, α, β, δ)*

I. Initialize wolf population by considering previously defined boundaries.
II. Estimate objective function for each wolf.
III. Select the best three fittest wolves and assign them as "α", "β" and "δ".
IV. Modify the position of the rest "w wolves" by employing Equations from 10 to 16.
V. Modify parameter "b", and vectors $\vec{R}$ **and** $\vec{C}$.
VI. Move back to "*step II*" as the required progress is not obtained.
VII. Yield the position of "α" as the finest approximation to the optimum solution.

## 3.3 Grey Wolf Optimizer for AE

One of the main contribution of this paper to the field is to employ GWO algorithm to train Auto-encoders in an unsupervised strategy. The critical problem here is to represent learning as fitness function. Essentially, during the training, it is aimed to obtain a set of values for weight and bias parameters so as to reduce the error rate close to zero. The parameters, representing the weights and basis, are formed as vector so to be accepted by GWO algorithm, as illustrated in Eq. 17.

$$\vec{V}(\vec{w}, \vec{b}) = (\vec{w_{11}}, \vec{w_{12}}, \ldots \vec{w_{ij}}, \vec{b_1}, \vec{b_2}, \ldots \vec{b_h}, \lambda, \beta) \tag{17}$$

Here, $w_{ij}$ is the linking weight from the $i_{th}$ node to the $j_{th}$ node, while $b_j$ represents the bias of the $j_{th}$ hidden node. The objective function of our problem represented in (5). The aim is to minimize the cost function (18). The training model of Auto-encoders can be expressed by considering the cost function and GWO algorithm as follows:

$$E = \min_{V}(E) \tag{18}$$

As aforementioned, our goal is to minimize the E value by estimating most suitable weights and bias values, as well as the control parameters, namely, λ and β. Figure 1 illustrates to the main procedure of training an auto-encoder by employing GWO algorithm. The GWO algorithm is applied to auto-encoders of a SAEs model. The GWO algorithm iteratively updates the weights, biases and control parameters so as to minimize E, cost function, of all training examples.
The weights/biases and control parameters are enhanced during each iteration and in high probability allows auto-encoders to extract better features from the input data during the training phase. Due to the stochastic behavior of metaheuristics approach, it is not guarantee to estimate the most optimum parameters for a dataset. However, by performing enough number of iterations the GWO eventually converges and approximate the optimum solution.

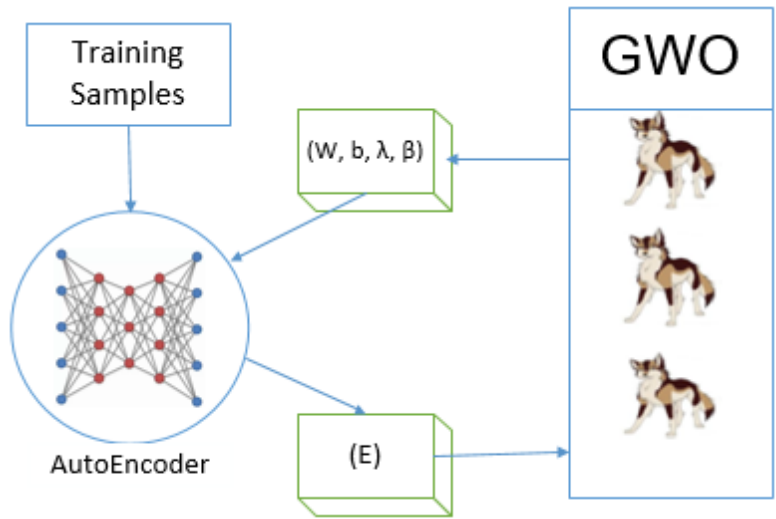

**Figure 1:** GWO Algorithm used to train AEs by aiming to minimize Cost Function

## 4. Experimental Results and Discussion

The system is implemented by employing open source frameworks supporting Python language and tested with six popular medical datasets, presented in Table 1, so as to evaluate the overall performance of the proposed system. These datasets are, namely, Leukemia, Colon, Prostate, Lung, Gene Expression Cancer RNA-Seq and Leukemia 2 . While 70% of them are used for training, 30% is used for testing. The mutual property between these datasets that all of them have high dimensional feature, presenting complex behaviour with low number of instances.

**Table 1:** Details of Datasets

| Dataset Name | Year | Features | Instances | Classes |
|---|---|---|---|---|
| Leukemia 1 [28] | 1999 | 7129 | 72 | 3 |
| Colon [29] | 1999 | 2000 | 62 | 2 |
| Prostate [30] | 2002 | 12,600 | 102 | 2 |
| Lung [31] | 2002 | 7129 | 96 | 2 |
| Gene Expression Cancer RNA-Seq [43] | 2013 | 20531 | 801 | 5 |
| Leukemia 2 [44] | 2019 | 22284 | 64 | 5 |

The GWO is applied to train the auto-encoders of a SSAE based Deep learning framework to extract high level features in minimum time to yield high classification accuracy. As it has been preliminary investigated, the classical training techniques tend to remain inefficient while attempting train these datasets.

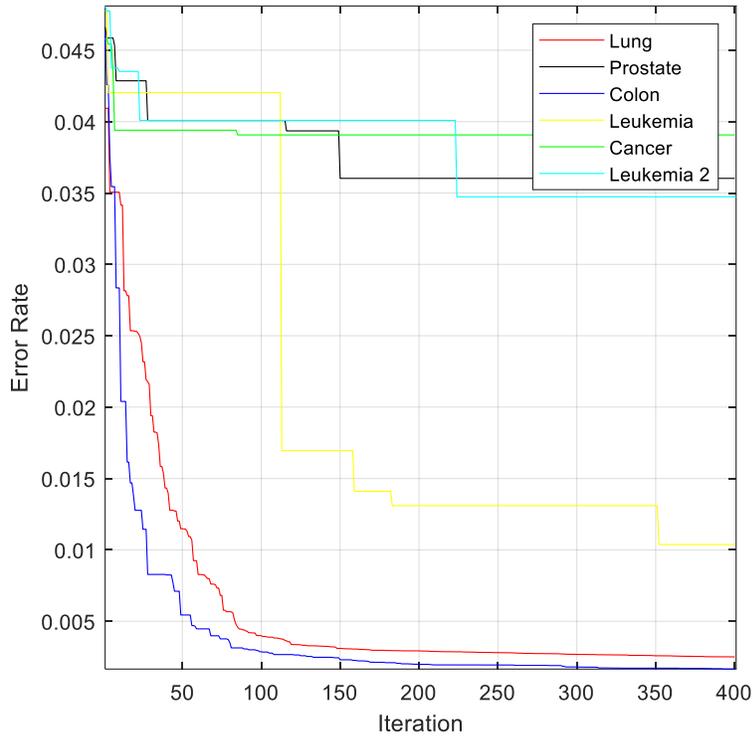

**Figure 2:** Variation of the error rate across iterations for the first AE for different datasets

The SSAE framework, detailed in Section 3.1, is employed to train the datasets given in Table 1. This framework essentially, transforms an unsupervised learning architecture into a supervised learning architecture, the details can be seen in [24]. The optimization procedure for each auto-encoder starts by initializing biases, weights and control parameters in the domain of [−20, 20] for each dataset randomly and individually. The first stage of the training is to obtain features by employing the first auto-encoder. The variation or the error rate across iterations for the first AE regarding each dataset during the training process, detailed in Table 1, illustrated in Figure 2. The output features of first auto-encoder, representing the best features with minimum error rate, are passed to the second auto-encoder.

The same optimizing procedure repeated with the second auto-encoder as well. The training procedure of each datasets for this AE is illustrated in Figure 3. Finally, the output of second auto-encoder is linked to the SoftMax layer.

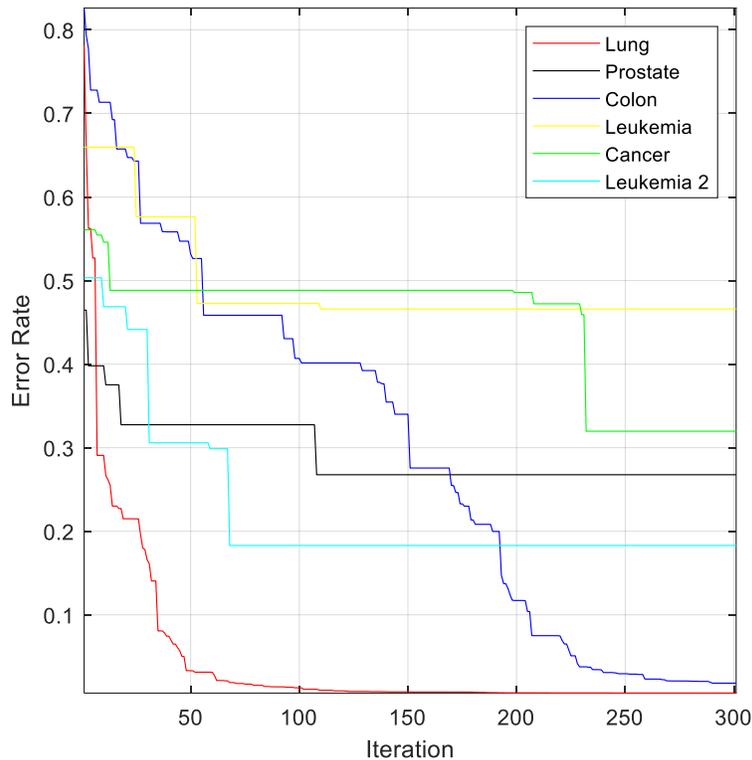

**Figure 3:** Variation of the error rate across iterations for the second AE for different datasets

The SoftMax, on the other hand is trained in a supervised manner to categorize the features into appropriate classes. Consequently, the two AEs and the SoftMax layers are stacked and trained in supervised technique by using labelled data.

As aforementioned, the proposed framework is validated by using four gene expression datasets, namely, Leukemia, Colon, Prostate, and Lung. Eexperiments are repeated five times for each dataset separately. The average accuracy is calculated and shared within this study, as well as, data are selected randomly for each experiment to avoid overfitting.

Overall, Leukemia dataset, consisting from 7129 features and 72 instances with 3 classes, the structure of the model is designed as "681-437-3. This model presents 99.02% accuracy within the given dataset. Furthermore, Colon cancer dataset involves 62 samples, each having 2000 features and is classified into 2 classes. The structure of the model is designed as 1631-1381-2, and performs 92.29% accuracy. Moreover, Prostate dataset includes 102 patients with 12,600 features for each case within 2 classes. The structure of the model is designed as "10201-6073-2", and yields 99.02% accuracy correspondingly. Lung dataset, on the other hand, consists from 96 patients, 7129 features for each case and 2 classes. The structure of the module is designed as 6320-4303-2, and produces 100% accuracy. Gene Expression Cancer RNA-Seq dataset, involving 20531 features for 801 instances with 5 classes. The structure of the model for this dataset is designed as "9324-3932-5", and it performed 99.70% accuracy. Finally, Leukemia 2 dataset, consisting from 22284 features and 64 instances with 5 classes, is employed to validate the performance of the system. The structure of the model for this dataset is designed as "1191-4321-5" and achieved to present 98.99 % accuracy.On the other hand, the accuracy of the model calculated by using the equation (18):

$$ACC = \frac{TP + TN}{TP + TN + FN + FP}$$

(8)

ACC metric represents the accuracy parameter, referring closeness of the measurements to a specific value. Here, TP presents the true positive, TN representes the true negative, FN denotes false negative and the FP is for the false positive values.

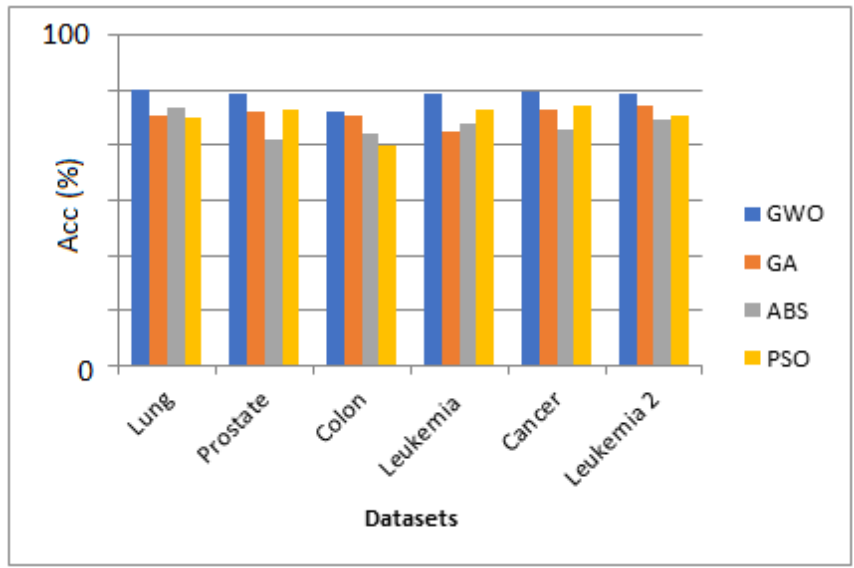

**Figure 4:** Comparison between several optimization algorithms on the same datasets

In order to reveal the advantages of the GWO algorithm to train AEs, it has been compared with the popular meta-heuristic algorithms, namely, "GA", "PSO", "ABC". These algorithms are also applied to train the framework, besides the trained framework is tested with aforementioned datasets under the same experimental conditions. Figure 4 presents and proves the superiority of GWO algorithm over other popular algorithms for this problem.  As well as, Figure 5 presents computational performance of these algorithms under the same experimental conditions. It is also proved that the GWO algorithm is able to complete the training procedure faster than other three popular optimization algorithms.

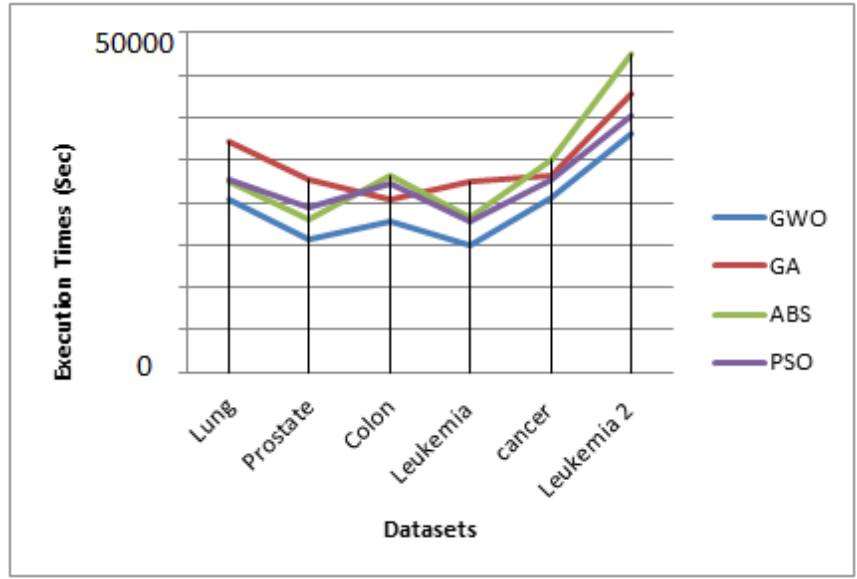

**Figure 5:** Comparison between runtimes of GWO, PSO, GA, and ABC.

*4 Discussion*

According to this inclusive study and experimental results there are several notes, revealing the superiority of GWO over other leading metaheuristic algorithms, listed below:

- The GWO algorithm provides high local optima avoidance mechanism, this can be noted from mathematical model of algorithm, devoting half of iteration in exploration of the search space and leads to explore various AEs structures during optimization procedure.
- The agent is forced randomly to take steps towards the prey by considering "C" parameter in GWO. This parameter essentially allows to avoid the local minima recession during the exploration process. Fortunately, the AEs training process requires high level local optima avoidance mechanism.
- Results reveal that the training of AEs with PSO and ABC algorithms presents deprived results. That is probably they do not have a mechanism provided by "C" parameter of GWO algorithm, allowing major unexpected maneuverers within the search.
- GA presents high accuracies compared with PSO and ABC algorithm. However, due to the characteristics of the GA, it requires longer execution time to approximate the optimal solution.
- The GWO-based trainer has ability to converge quickly to the global optimum in different datasets because of having high exploitative behavior.
- The GWO works efficiently with small and high dimensional dataset in contrast to the gradient-based algorithms that tend to fail with high dimensional datasets.

Table 2 presents comparison results between the proposed framework and the state-of-the-art studies using the same datasets for validation. The major drawbacks of previous studies are to rely on traditional techniques, most of which do not provide successful results with high dimensional small datasets. As aforementioned, the results prove the superiority of the proposed studies over other studies based on the classification problem using Gene expression datasets. As it can be seen from the corresponding Table, the previous studies do not conduct experiments with the given dataset is illustrated with "-" sign. It should be noted that most state-of-the-art studies do not carry out experiment with especially "Cancer" and "Lukemia 2" datasets.

**Table 2:** Accuracy (%) comparison between the proposed system and the state-of-the-art studies

| Ref | Methods | Lung (%) | Prostate | Colon | Leukemia | Cancer | leukemia 2 |
|---|---|---|---|---|---|---|---|
| [19] | "MIMAGA + ELM" | 97.80 | 97.69 | 89.09 | 97.62 | - | - |
| [19] | "ReliefF + ELM" | 54.23 | 61.15 | 68.18 | 68.18 | - | - |
| [19] | "SFS + ELM" | 89.57 | 86.28 | 70.63 | 96.88 | - | - |
| [19] | "MIM + ELM" | 79.52 | 88.67 | 68.17 | 76.83 | - | - |
| [20] | "KNN" | 99.14 | - | - | 81.92 | - | - |
| [20] | "ReliefF" | 99.14 | - | - | 84.79 | - | - |
| [20] | "MReliefF" | 99.29 | - | - | 85.54 | - | - |
| [20] | "ROS+ReliefF" | 99.27 | - | - | 85.33 | - | - |
| [20] | "RUS+ReliefF" | 98.93 | - | - | 0.8725 | - | - |
| [20] | "imRelief" | 99.28 | - | - | 86.17 | - | - |
| [21] | "DFS+SVM" | 98.61 | - | 87.09 | 98.61 | - | - |
| [22] | "IG-SVM" | 100 | 96.08 | 90.32 | 98.61 | - | - |
| [22] | "GR-SVM" | 98.96 | 93.14 | 85.48 | 94.44 | - | - |
| [22] | "ReliefF-SVM" | 98.96 | 91.18 | 87.10 | 97.22 | - | - |
| [22] | "Cor-SVM" | 98.96 | 93.14 | 87.10 | 97.22 | - | - |
| [46] | "SVM" | - | - | - | - | 95.00 | - |
| [47] | "Feature subset-based ensemble method" | - | - | - | - | 98.60 | - |
| [48] | "Random Forest" | - | - | - | - | 98.77 | - |
| [49] | "Grouping Genetic Algorithm (GGA)" | | | | | 98.81 | - |
| [50] | "t-SNE+PCA+ZeroR" | - | - | - | - | - | 41.0 |
| [50] | "t-SNE+PCA+SVM" | - | - | - | - | - | 98.0 |
| [50] | "t-SNE+PCA+MLP" | - | - | - | - | - | 94.0 |
| [50] | "t-SNE+PCA+DT" | - | - | - | - | - | 89.0 |
| [50] | "t-SNE+PCA+NB" | - | - | - | - | - | 89.0 |
| [50] | "t-SNE+PCA+RF" | - | - | - | - | - | 98.0 |
| [50] | "t-SNE+PCA+HC" | - | - | - | - | - | 41.0 |
| [50] | "t-SNE+PCA+KNN" | - | - | - | - | - | 89.0 |
| [50] | "t-SNE+PCA+K-mean" | - | - | - | - | - | 67.0 |
| [50] | Relief algorithm+ wrapper method + GA+ SVM | - | - | - | - | - | 95.31 |
| [50] | Relief algorithm+ wrapper method + GA+ Naïve Bayes | - | - | - | - | - | 95.31 |
| [50] | Relief algorithm+ wrapper method + GA+ Decision Tree | - | - | - | - | - | 95.31 |
| [50] | Relief algorithm+ wrapper method + GA+ K nearest | - | - | - | - | - | 96.88 |

| | DSAE based GWO | 100 | 99.02 | 92.29 | 99.02 | 99.70 | 98.99 |

## 5. Conclusion

This paper proposes a novel gene expression data classification framework based on deep learning. The main contribution of this study is to adapt Grey Wolf Optimizer (GWO) algorithm to train AEs by considering an unsupervised learning procedure. This meta-heuristic algorithm enables the SSAE based algorithm to extract better features that, in essence, allows the framework to present remarkable results with popular biomedical datasets.

The GWO algorithm is also compared with comprehensive metaheuristic algorithms, namely, Genetic Algorithms (GA), Particle Swarm Optimization (PSO) and Artificial Bee Colony (ABC) under the same experimental conditions for the same datasets. The superiority of the GWO algorithm over its counterparts for this problem is also validated. Furthermore the proposed framework is validated by employing six benchmark datasets with respect to the gene expression field. The proposed framework also outperforms the state-of-the-art studies. Results lead us to clarify that the SAEs using GWO algorithm performs successfully even in the presence of small and high dimensional dataset in contrast to the gradient-based algorithms that are likely to fail with high dimensional datasets. These results encouraged authors to employ this framework on biomedical data classification problems, as well as for other wide-ranging problems, each of which can only provide small datasets with high dimensional complex data.

## Conflicts of Interest

The authors declare that there are no conflicts of interest regarding the publication of this paper.